\documentclass{article}

\usepackage[preprint, nonatbib]{neurips_2023}

\usepackage[utf8]{inputenc} 
\usepackage[T1]{fontenc}    
\usepackage{hyperref}       
\usepackage{url}            
\usepackage{booktabs}       
\usepackage{amsfonts}       
\usepackage{nicefrac}       
\usepackage{microtype}      
\usepackage{xcolor}         
\usepackage{graphicx}
\usepackage{comment}
\usepackage{amsmath}
\usepackage{adjustbox}
\usepackage{multirow}

\title{Predicting the Algorithm Tags and Difficulty \\ for Competitive Programming Problems}

\author{Juntae Kim$^{1}$ \; Eunjung Cho$^{2}$ \; \stepcounter{footnote}Dongbin Na$^{3}$\thanks{Correspondence to dongbinna@postech.ac.kr} \\ $^1$Yonsei University \; $^2$Inha University \; $^3$POSTECH}

\begin{document}

\maketitle

\begin{abstract}

The recent program development industries have required problem-solving abilities for engineers, especially application developers.
However, AI-based education systems to help solve computer algorithm problems have not yet attracted attention, while most big tech companies require the ability to solve algorithm problems including Google, Meta, and Amazon.
The most useful guide to solving algorithm problems might be guessing the category (tag) of the facing problems.
Therefore, our study addresses the task of predicting the algorithm tag as a useful tool for engineers and developers.
Moreover, we also consider predicting the difficulty levels of algorithm problems, which can be used as useful guidance to calculate the required time to solve that problem.
In this paper, we present a real-world algorithm problem multi-task dataset, \textbf{AMT}, by mainly collecting problem samples from the most famous and large competitive programming website Codeforces.
To the best of our knowledge, our proposed dataset is the most large-scale dataset for predicting algorithm tags compared to previous studies.
Moreover, our work is the first to address predicting the difficulty levels of algorithm problems.
We present a deep learning-based novel method for simultaneously predicting algorithm tags and the difficulty levels of an algorithm problem given.
All datasets and source codes are available at \textcolor{blue}{\url{https://github.com/sronger/PSG_Predicting_Algorithm_Tags_and_Difficulty}}.
\end{abstract}

\section{Introduction}

To solve a given algorithm problem, in general, the developer guesses the intention of the problem and classifies the algorithm tag of the problem after reading the problem description.
Then, the developer writes a source code for solving the algorithm problem.
From this perspective, a problem-solving guide (PSG) is a useful tool for learners and engineers who are facing algorithm problems.
For example, predicting algorithm tags properly for a given problem description can provide useful direction to understand the problem for the participants.
Moreover, the order of problems to solve also does matter because we may not have enough time to solve all problems.
Thus, we note that predicting the difficulty level is also informative in deciding the order to solve problems.
In this paper, we introduce an AI-based problem-solving guide (PSG) as a useful tool for programmers facing an algorithm problem.
Given algorithm problems, our PSG is a multi-task solution providing simultaneously (1) a predicted algorithm tag, and (2) a predicted difficulty level of the problem.
For educational purposes, our proposed method can be used to reduce effectively the time for users to understand and solve various algorithm problems.

Recent work has shown that the deep learning-based classifier can be used for predicting the algorithm tags of a problem given~\cite{pwp}.
However, they have a limitation in that their method is able to only predict an algorithm tag for a problem and shows poor classification accuracy.
For a generalized problem-solving guide, we should design proper architectures that predict the tags and difficulty of the problem properly by understanding the intent of the problem comprehensively.
Especially, the algorithm problem set consists of long sentence texts.
We note that the deep learning models based on recurrent neural networks~\cite{lstm,bigbird,longformer} are difficult to recognize these long sentences.
In this work, we utilize a useful deep-learning architecture to effectively address these long-sequence texts.
We adopt transformer-based large language models~\cite{transformer,bert,electra} and show the recent transformer architectures that address long sequences are useful for solving our task~\cite{longformer,bigbird}.

We have also analyzed various programming problems in the broadly used competitive programming platform, Codeforces.
On this website, the categories of problems are labeled by algorithm experts, and the difficulty levels are determined by the results of the competition to which the problem belongs except in a few exceptional cases.
Predicting the categories of the problem is a well-defined multi-label problem~\cite{multi1, multi2, multi3, multi4, multi5} and predicting the problem's difficulty level can be seen as the ordinal classification problem~\cite{meanvariance, ordinal, simpleordinal, ordinal1, ordinal2, ordinal3}.
Therefore, we consider this problem a multi-task problem that jointly solves two tasks and also provide a new dataset, \textbf{AMT}.
We demonstrate that our proposed method shows superior classification performance compared to the previous SOTA work~\cite{pwp}.
To the best of our knowledge, we are the first to adopt the multi-task approach to provide useful applications for real-world developers, which simultaneously predicts the tags and difficulties of an algorithm problem.
Moreover, we additionally provide baseline source code templates for the various algorithm tags, which can reduce the time effectively for users to solve the algorithm problems.

\begin{figure}
\centering
\includegraphics[width=1.0\linewidth]{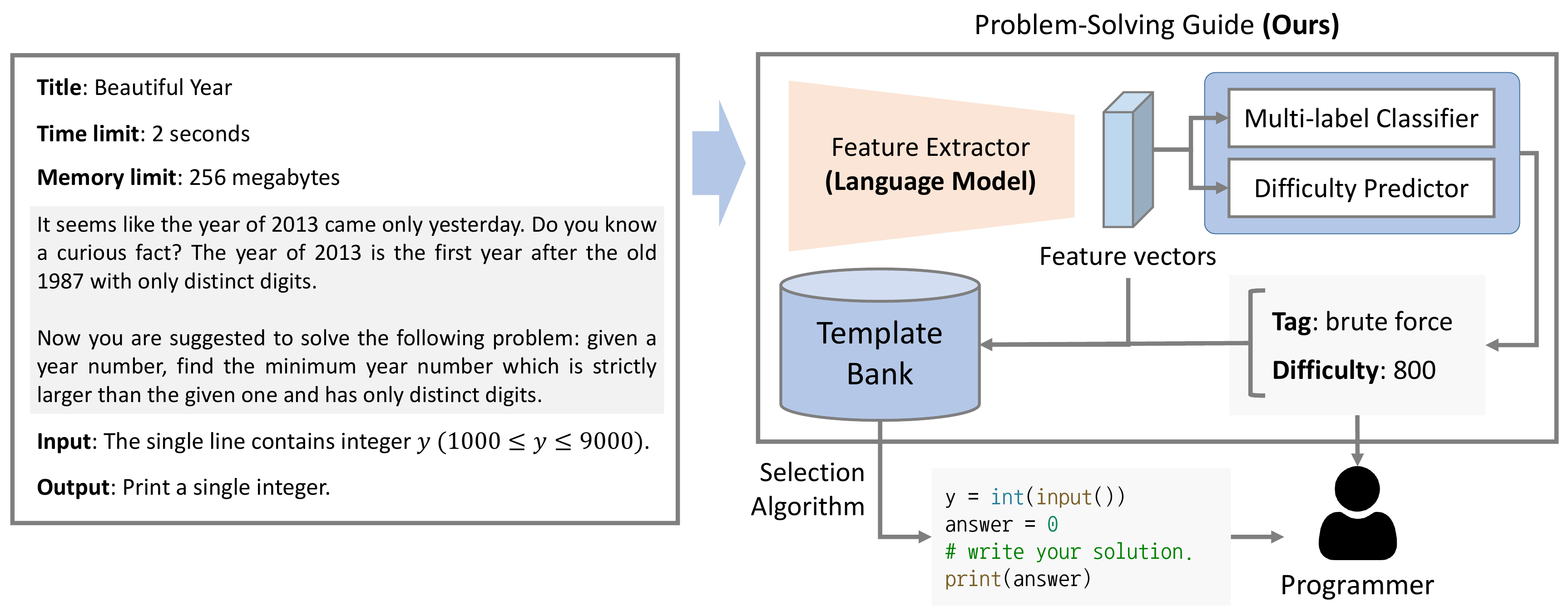}
\caption{\label{fig:overview}Our proposed method, problem-solving guide (PSG) predicts the tags (categories) and the difficulty (required time) of an algorithm problem simultaneously.}
\end{figure}

\section{Background and Related Work}

\subsection{Predicting Problem Tags}

A recent study has proposed a new research area PMP (Programming Word Problems) and presented a dataset for the research purpose of predicting algorithm tags~\cite{pwp}.
In their work, the authors utilize 4,019 problems and more than 10 algorithm tags.
They have demonstrated that the CNN-based classifier can achieve near-human performance for the task of predicting the algorithm tags~\cite{pwp}.
However, their adopted architectures do not address the long sequences effectively.
The general algorithm problems consist of a lot of words whose size is more than 1,000 in the problem description.
However, we have found that the simple CNN architecture with fixed kernel size might not capture the global representations comprehensively. 
To remedy this issue, we adopt the recent transformer-based architectures~\cite{transformer,bert,bigbird} that are relatively immune to long sequences of the problem description.
Moreover, we extend the number of algorithm problems to construct large-scale datasets.

\subsection{Multi-Task Solution Using Deep Learning}

We consider our task as a multi-task problem that simultaneously addresses (1) multi-label classification and (2) ordinal-class classification.
Some related studies have presented joint learning methods for simultaneously training multiple tasks in various research fields~\cite{multitaskoverview, multitask, multitasksurvey, multi12, multi13}.
The multi-task learning approach can reduce memory complexity while nearly maintaining the original classification performance of individual single-task models~\cite{multitask, multitaskoverview, multi10, multi11}.

\begin{table}
\caption{\label{tab:label_distribution}The data distribution of algorithm tags. This table shows the number of algorithm problems for each tag category in our presented dataset. We consider the most frequent 20 categories.}
\centering
\begin{adjustbox}{width=14.0cm,center}
\begin{tabular}{|l|r|l|r|l|r|}
\hline
\multicolumn{6}{|c|}{Top-20 Frequent Categories} \\ \hline
Labels & \# of problems & Labels & \# of problems & Labels & \# of problems \\ \hline
Implementation & 2394 & Sortings & 869 & Bitmasks & 459 \\
Math & 2363 & Binary Search & 862 & Two Pointers & 438 \\
Greedy & 2302 & DFS and Similar & 776 & Geometry & 344 \\
DP & 1732 & Trees & 663 & DSU & 292 \\
Data Structures & 1429 & Strings & 617 & Shortest Paths & 231 \\
Brute Force & 1370 & Number Theory & 613 & Divide and Conquer &  227\\
Graphs & 890 & Combinatorics & 544 & & \\
\hline
\end{tabular}
\end{adjustbox}
\end{table}

\begin{table}[]
\caption{\label{tab:difficulty_distribution}The difficulty level distribution of our proposed whole dataset. This table shows the number of algorithm problems according to the difficulty levels. CodeForces provides 28 different types of difficulty levels.}
\centering
\begin{adjustbox}{width=13.0cm,center}
\begin{tabular}{|c|c|c|c|c|c|c|c|c|c|c|}
\hline
\textbf{Difficulty}     & 800  & 900  & 1000 & 1100 & 1200 & 1300 & 1400 & 1500 & 1600 & 1700 \\ \hline
\textbf{\# of Problems} & 686  & 255  & 306  & 305  & 333  & 325  & 329 & 357 & 397 & 381  \\ \hline
\textbf{Difficulty}    & 1800 & 1900 & 2000 & 2100 & 2200 & 2300 & 2400 & 2500 & 2600 & 2700 \\ \hline
\textbf{\# of Problems} & 348  & 371  & 363  & 330 & 362  & 297  & 347  & 306  & 242  & 222  \\ \hline
\textbf{Difficulty}     & 2800 & 2900 & 3000 & 3100 & 3200 & 3300 & 3400 & 3500 & & \\ \hline
\textbf{\# of Problems} & 177 & 165  & 137  & 107  & 105  & 86   & 63   & 112 & & \\ \hline
\end{tabular}
\end{adjustbox}
\end{table}

\section{Proposed Methods}

Our method solves the \textbf{multi-label classification} problem because each algorithm problem can belong to one or more labels simultaneously.
For example, a competitive programming problem requires the idea of \textit{greedy}, \textit{sorting}, and \textit{dynamic programming} simultaneously.
Our proposed method also predicts the difficulty of the problems.
To the best of our knowledge, we are the first to address the \textbf{ordinal-class classification} problem for predicting the degree of difficulty of algorithm problems.
We note that predicting the degree of difficulty of algorithm problems is also crucial for developers in that the difficulty level can be interpreted as the required time to solve that algorithm problem.

\subsection{Problem Definition}

We define a function $F:\mathcal{X} \rightarrow \mathcal{Z}$ as a feature extractor that extracts representations given a text $x$ and maps $x$ into an embedding space $\mathcal{Z}$.
Then, we use a classification head $H:\mathcal{Z} \rightarrow \mathcal{Y}$ on the top of the feature extractor $F$.
Our proposed framework is designed to solve multiple tasks.
Specifically, our model solves two kinds of problems (1) multi-label classification and (2) ordinal-class classification simultaneously.
Thus, we design deep neural networks for solving these two tasks using two loss functions $l_1$ and $l_2$ jointly.

$$ \mathbb{E}_{(x, y, d) \in \mathcal{D}_{train}}[l_1(H_1(z), y)) + \lambda \cdot l_2(H_2(z), d))] $$

where $z = F(x)$ and $l_1$ denotes a binary cross-entropy loss for the problem category $y$.
$\mathcal{D}_{train}$ denotes the train data distribution.
The second loss $l_2$ is designed for the ordinal classification task.
Thus, we can simply adopt the cross-entropy loss for $l_2$.
The $d$ denotes a problem difficulty level and the $\lambda$ is a scale factor for weighting two tasks differently.
Our PSG adopts the two-head network architecture.
First, we extract a feature representation vector $z$ and forward this vector into two classification heads, multi-label classifier head $H_{1}$ and ordinal-class classifier head $H_{2}$.
During the training time, the data $x$ and $y$ are picked from the train data distribution $\mathcal{D}_{train}$.
After training time, we test the trained model on the test dataset $\mathcal{D}_{test}$ that is different from $\mathcal{D}_{train}$ in the inference time.

In the multi-label classification tasks, the classification model can classify multiple labels simultaneously, performing binary classification per each label.
Thus, we adopt the multi-label classification approach for solving the algorithm tag prediction task in this work.
Given a problem description, our model outputs the probability for each possible algorithm tag label.
Thus, we can adopt the binary cross-entropy (BCE) loss function.
For the following equation, $y_k$ is set as 1 where the text data $x$ belongs to the $k$-th class and the value of $y_k$ is $0$ in the case that $x$ does not belong to the $k$-th category (tag).
Therefore, we can calculate the full loss value over all possible categories where the number of categories is $K$ for the multi-label classification and $y \in \mathbb{R}^K$ denotes the true label according to $x$.
We have shown that this simple BCE loss is well suitable for our multi-label classification problem where a detection network $\psi(\cdot)$ that informs users of whether a text data belongs to the $\mathcal{D}^{k}$ that is data distribution of $k$-th algorithm category (tag), thus, $\psi\left(x\right) = 0$ if $x \notin \mathcal{D}^{k}$.

\vspace{-0.5cm}
\begin{align}
\mathcal{L}^{tag}(\psi(x),y) = -\frac{1}{K}\sum_{k}^{K}{y_k \cdot log(\psi(x)) + (1 - y_k) \cdot log(1 - \psi(x))}
\end{align}

\subsection{Proposed Datasets and Architectures}

To construct our dataset, \textbf{AMT}, we have mainly collected algorithm problems from CodeForces.
We have excepted a problem if the problem has no tag information.
The total number of collected problems is 7,976.
First, we consider the top 20 frequent algorithm tags as ground-truth labels, which account for most problems.
We represent the number of programming problems for each problem tag of our proposed dataset in Table~\ref{tab:label_distribution}.
Secondly, our proposed dataset also contains the difficulty information for each problem.
The smaller value denotes the easier problem in Table~\ref{tab:difficulty_distribution}.
In our proposed dataset, the difficulty level is calculated based on their own rating system of Codeforces.
To implement the multi-task deep learning model, our method adopts BERT-based~\cite{bert,bigbird,transformer} feature extractor and two different classification head networks.
With extensive experiments, we have found that the BigBird~\cite{bigbird} architecture shows the best performance for solving our multi-task problem.
Therefore, we report the performance of BigBird as the main result in the experiment section.

\begin{table}[]

\caption{\label{tab:main_results}Performance on the test dataset. The symbol $\uparrow$ indicates larger values are better. We have reported the best performance of each method by finding the best learning rate using a grid search. In conclusion, we use the learning rate of 5e-6 for BigBird models solving each single task. Moreover, we also use the learning rate of 5e-6 for our proposed multi-task solver PSG. We note that the hyper-parameter $\lambda$ is crucial to obtain improved classification performance. For the tag prediction, the AUROC and F1-Macro indicate the average value over all the categories.
We also report the performance of some baseline methods of the previous work.
}

\centering
\begin{adjustbox}{width=14.0cm,center}
\begin{tabular}{|c|c|c|c|c|c|c|c|}
\hline
\multirow{3}{*}{\textbf{Architectures}} & \multirow{3}{*}{\textbf{$\lambda$}} & \multicolumn{4}{|c|}{\textbf{Rating Prediction $\mathcal{T}_{1}$}} & \multicolumn{2}{|c|}{\textbf{Tag Prediction $\mathcal{T}_{2}$}} \\
& & Accuracy & CS ($\theta$=3)~\cite{meanvariance} & CS ($\theta$=5)~\cite{meanvariance} & MAE & AUROC & F1-Macro \\
& & $\uparrow$ & $\uparrow$ & $\uparrow$ & $\downarrow$ & $\uparrow$ & $\uparrow$ \\ \hline
SVM BoW + TF-IDF~\cite{pwp} & N/A & N/A & N/A & N/A & N/A & 79.26 & 40.33 \\
CNN Ensemble TWE~\cite{pwp} & N/A & N/A & N/A & N/A & N/A & 58.12 & 20.14  \\ \hline

XGBoost & N/A & N/A & N/A & N/A & N/A & 73.47 & \textbf{43.09}  \\ 
CatBoost & N/A & N/A & N/A & N/A & N/A & 74.39 & 42.91  \\ 
LightGBM & N/A & N/A & N/A & N/A & N/A & 74.52 & 42.94  \\ 
Gradient Boosting Machine & N/A & N/A & N/A & N/A & N/A & 73.21 & 40.99  \\ \hline

BigBird-based Single Model for $\mathcal{T}_{1}$ (\textbf{Ours}) & N/A & \textbf{11.24} & \textbf{23.71} & \textbf{34.54} & \textbf{4.55} & N/A & N/A \\
BigBird-based Single Model for $\mathcal{T}_{2}$ (\textbf{Ours}) & N/A & N/A & N/A & N/A & N/A & \textbf{80.70} & 42.78 \\ \hline

Multi-Task PSG (\textbf{Ours}) & 1 & 8.35 & 19.07 & 32.68 & 4.74 & 69.08 & 25.16 \\
Multi-Task PSG (\textbf{Ours}) & 10 & 10.10 & 20.41 & 33.71 & 4.79 & 79.12 & 41.08 \\  
Multi-Task PSG (\textbf{Ours}) & 100 & 8.76 & 15.46 & 23.20 & 7.09 & 79.59 & 41.63 \\ \hline

\end{tabular}
\end{adjustbox}
\end{table}

\section{Experimental Results}

We have extensively experimented with various text classification methods to validate the effectiveness of our proposed dataset. We also construct a smaller version of our AMT dataset, \textit{AMT10}, that only considers the main 10 categories for experiments. A recent work~\cite{pwp} has applied various deep learning-based methods for predicting algorithm tags and has demonstrated CNN architectures~\cite{cnn1,cnn2} could show improved performance.
However, we observe that the recently proposed large-scale transformer architectures~\cite{bigbird,bert} can show better classification performance compared to the reported classification performance of previous methods as shown in Table~\ref{tab:main_results}.
With extensive experiments, we have found that the BERT-based architectures can be greatly useful for our task, especially the BigBird can comprehensively process the long embedding tokens and relatively well recognize the implicit feature representations of an algorithm problem.
In conclusion, our proposed method for multi-task learning, PSG, results in a competitive classification performance comprehensively.
We note that the number of parameters of PSG is smaller by approximately 2 times compared to the parameter size of the combination of two single-task models.
This memory efficiency comes from the property of the multi-task models that we can extract feature representations by forwarding the input texts into the feature extractor network $F(\cdot)$ only once.

\begin{figure}[h]
\centering
\begin{tabular}{cc}
\hspace{-0.8cm}
\includegraphics[width=0.35\textwidth ]{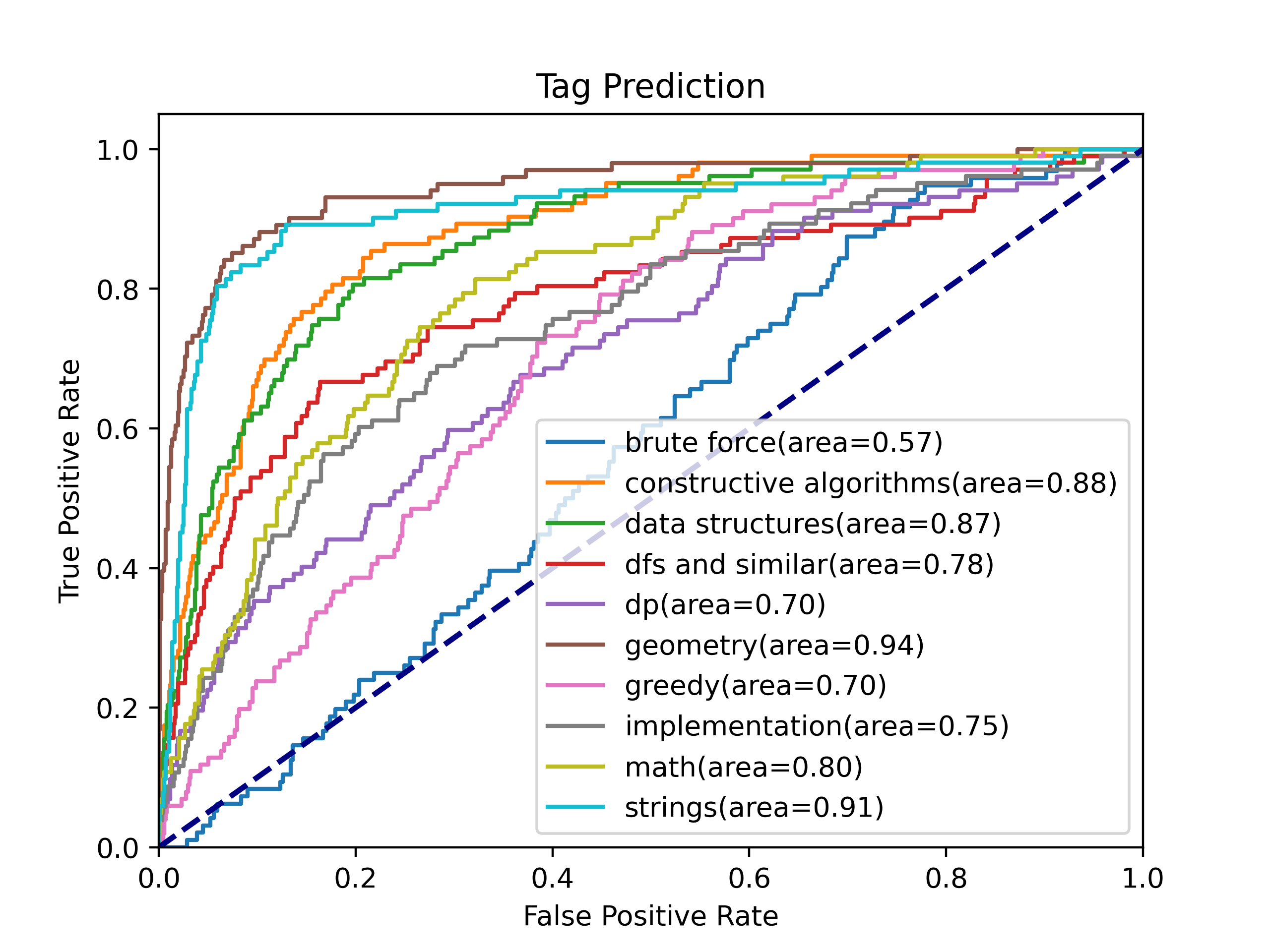}
\includegraphics[width=0.35\textwidth ]{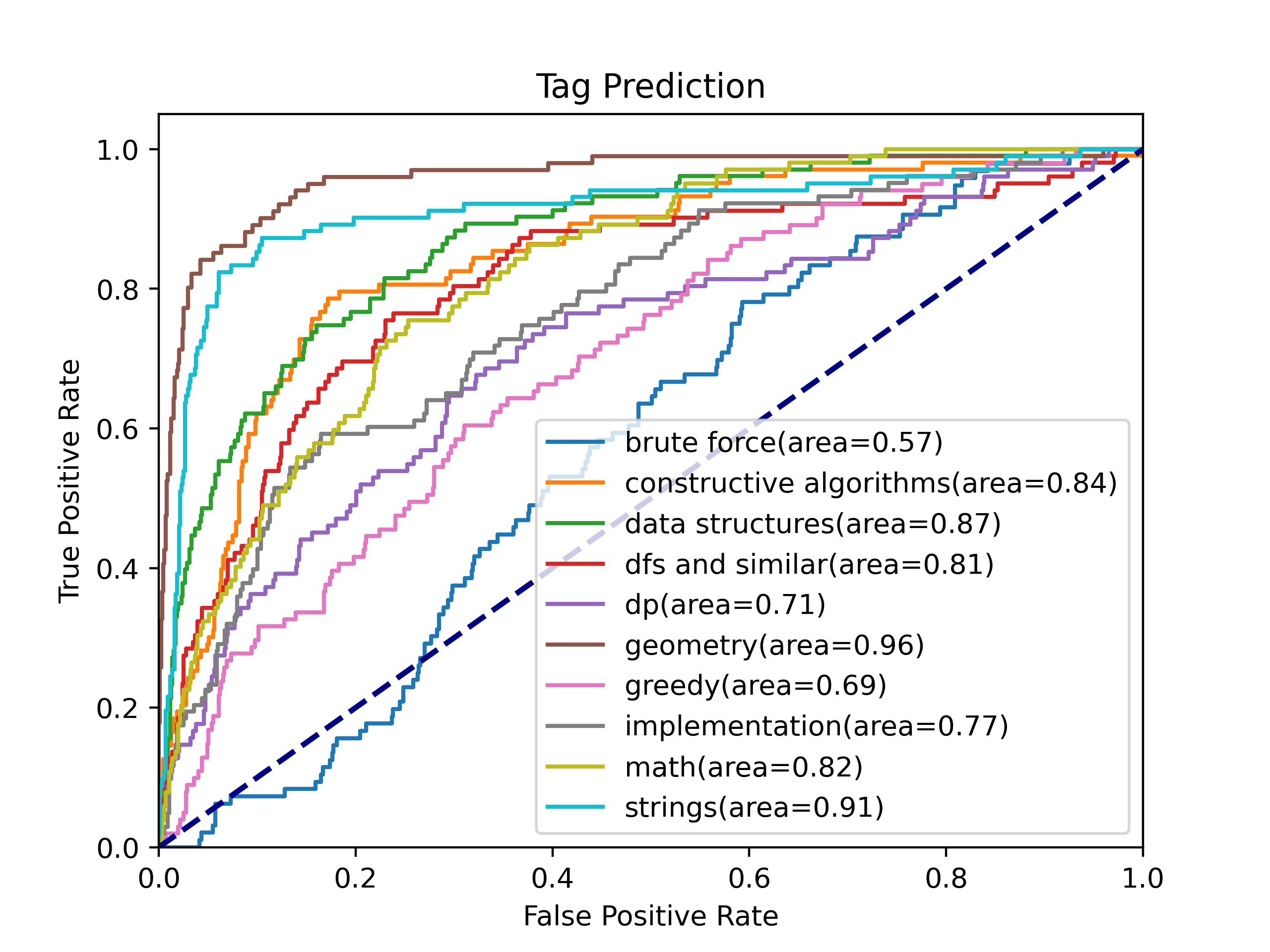}
\includegraphics[width=0.35\textwidth ]{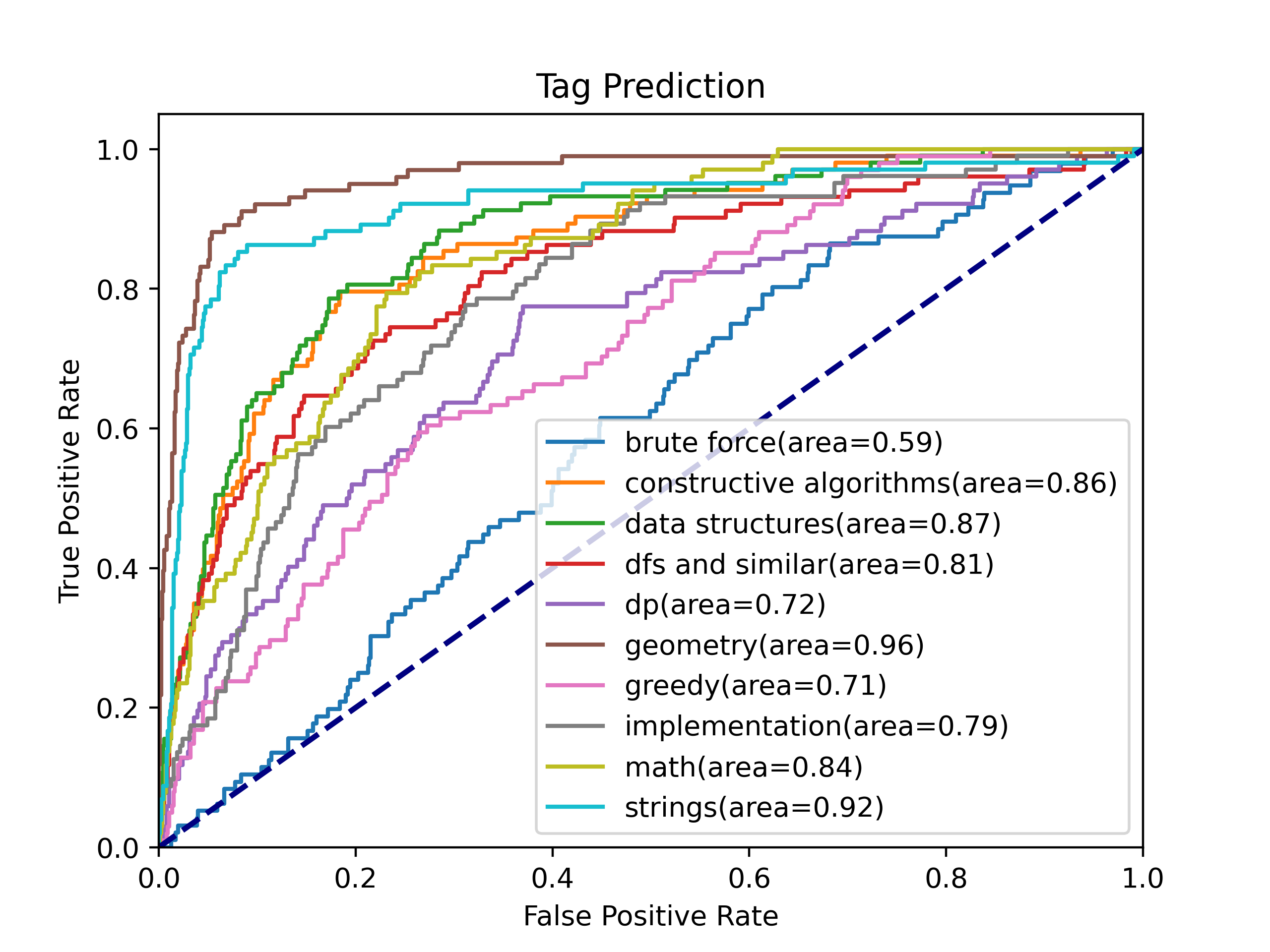} \quad \quad
\end{tabular}
\caption{\label{fig:figure_1}The ROC curves of various models, derived on the test dataset. The first figure represents the ROC curve of our PSG model trained with $\lambda=10$. The second figure shows the ROC curve of our PSG model trained with $\lambda=100$. The third figure shows the ROC curve of the single-task BigBird model to solve only the second task $\mathcal{T}_2$. Our multi-task learning method can obtain a competitive classification performance for the tag prediction task $\mathcal{T}_2$ compared to the single-task learning method while maintaining the ability to solve two different tasks.}
\end{figure}

\section{Conclusion}

In this work, we present a novel algorithm problem classification dataset, \textbf{AMT}, that contains about 8,000 algorithm problems and provides two kinds of annotations (the algorithm tag and difficulty level) for each problem.
To validate the effectiveness of the proposed dataset, we also train a variety of text classification models on the dataset and analyze their classification performance.
To solve the multi-task problem effectively, we introduce a novel multi-task approach, PSG, that simultaneously predicts the tag and difficulty level of an algorithm problem given.
In the experimental results, we demonstrate our proposed method shows significantly improved classification performance compared to the previously presented SOTA methods.
We provide all the source codes, datasets, and trained models publicly available.
We hope our proposed dataset and model architectures could contribute to the programming industries for educational purposes.

\bibliographystyle{plain}
\bibliography{neurips_2023}

\begin{thebibliography}{10}

\bibitem{pwp}
Vinayak Athavale, Aayush Naik, Rajas Vanjape, and Manish Shrivastava.
\newblock Predicting algorithm classes for programming word problems.
\newblock In {\em Conference on Empirical Methods in Natural Language Processing}, 2019.

\bibitem{longformer}
Iz~Beltagy, Matthew~E Peters, and Arman Cohan.
\newblock Longformer: The long-document transformer.
\newblock {\em arXiv preprint arXiv:2004.05150}, 2020.

\bibitem{multitask}
Dasol Choi, Jooyoung Song, Eunsun Lee, Jinwoo Seo, Heejune Park, and Dongbin Na.
\newblock Large-scale korean text dataset for classifying biased speech in real-world online services, 2023.

\bibitem{electra}
Kevin Clark, Minh-Thang Luong, Quoc~V Le, and Christopher~D Manning.
\newblock Electra: Pre-training text encoders as discriminators rather than generators.
\newblock {\em arXiv preprint arXiv:2003.10555}, 2020.

\bibitem{multitasksurvey}
Michael Crawshaw.
\newblock Multi-task learning with deep neural networks: A survey.
\newblock {\em arXiv preprint arXiv:2009.09796}, 2020.

\bibitem{ordinal3}
Krzysztof Dembczy{\'n}ski, Wojciech Kot{\l}owski, and Roman S{\l}owi{\'n}ski.
\newblock Ordinal classification with decision rules.
\newblock In {\em Mining Complex Data: ECML/PKDD 2007 Third International Workshop, MCD 2007, Warsaw, Poland, September 17-21, 2007, Revised Selected Papers 3}, pages 169--181. Springer, 2008.

\bibitem{bert}
Jacob Devlin, Ming-Wei Chang, Kenton Lee, and Kristina Toutanova.
\newblock Bert: Pre-training of deep bidirectional transformers for language understanding.
\newblock {\em arXiv preprint arXiv:1810.04805}, 2018.

\bibitem{multi11}
Daxiang Dong, Hua Wu, Wei He, Dianhai Yu, and Haifeng Wang.
\newblock Multi-task learning for multiple language translation.
\newblock In {\em Proceedings of the 53rd Annual Meeting of the Association for Computational Linguistics and the 7th International Joint Conference on Natural Language Processing (Volume 1: Long Papers)}, pages 1723--1732, 2015.

\bibitem{simpleordinal}
Eibe Frank and Mark Hall.
\newblock A simple approach to ordinal classification.
\newblock In {\em Machine Learning: ECML 2001: 12th European Conference on Machine Learning Freiburg, Germany, September 5--7, 2001 Proceedings 12}, pages 145--156. Springer, 2001.

\bibitem{ordinal1}
Lisa Gaudette and Nathalie Japkowicz.
\newblock Evaluation methods for ordinal classification.
\newblock In {\em Advances in Artificial Intelligence: 22nd Canadian Conference on Artificial Intelligence, Canadian AI 2009 Kelowna, Canada, May 25-27, 2009 Proceedings 22}, pages 207--210. Springer, 2009.

\bibitem{lstm}
Sepp Hochreiter and Jürgen Schmidhuber.
\newblock {Long Short-Term Memory}.
\newblock {\em Neural Computation}, 9(8):1735--1780, 11 1997.

\bibitem{multi5}
Ioannis Katakis, Grigorios Tsoumakas, and Ioannis Vlahavas.
\newblock Multilabel text classification for automated tag suggestion.
\newblock {\em ECML PKDD discovery challenge}, 75:2008, 2008.

\bibitem{cnn1}
Yoon Kim.
\newblock Convolutional neural networks for sentence classification.
\newblock {\em arXiv preprint arXiv:1408.5882}, 2014.

\bibitem{cnn2}
Siwei Lai, Liheng Xu, Kang Liu, and Jun Zhao.
\newblock Recurrent convolutional neural networks for text classification.
\newblock In {\em Proceedings of the AAAI conference on artificial intelligence}, volume~29, 2015.

\bibitem{ordinal2}
OI~Larichev and HM~Moshkovich.
\newblock An approach to ordinal classification problems.
\newblock {\em International Transactions in Operational Research}, 1(3):375--385, 1994.

\bibitem{multi10}
Xi~Lin, Hui-Ling Zhen, Zhenhua Li, Qing-Fu Zhang, and Sam Kwong.
\newblock Pareto multi-task learning.
\newblock {\em Advances in neural information processing systems}, 32, 2019.

\bibitem{multi3}
Jingzhou Liu, Wei-Cheng Chang, Yuexin Wu, and Yiming Yang.
\newblock Deep learning for extreme multi-label text classification.
\newblock In {\em Proceedings of the 40th international ACM SIGIR conference on research and development in information retrieval}, pages 115--124, 2017.

\bibitem{multi1}
Shuhua~Monica Liu and Jiun-Hung Chen.
\newblock A multi-label classification based approach for sentiment classification.
\newblock {\em Expert Systems with Applications}, 42(3):1083--1093, 2015.

\bibitem{ordinal}
Yanzhu Liu, Adams Wai~Kin Kong, and Chi~Keong Goh.
\newblock A constrained deep neural network for ordinal regression.
\newblock In {\em Proceedings of the IEEE conference on computer vision and pattern recognition}, pages 831--839, 2018.

\bibitem{multi2}
Jinseok Nam, Jungi Kim, Eneldo Loza~Menc{\'\i}a, Iryna Gurevych, and Johannes F{\"u}rnkranz.
\newblock Large-scale multi-label text classification—revisiting neural networks.
\newblock In {\em Machine Learning and Knowledge Discovery in Databases: European Conference, ECML PKDD 2014, Nancy, France, September 15-19, 2014. Proceedings, Part II 14}, pages 437--452. Springer, 2014.

\bibitem{meanvariance}
Hongyu Pan, Hu~Han, Shiguang Shan, and Xilin Chen.
\newblock Mean-variance loss for deep age estimation from a face.
\newblock In {\em Proceedings of the IEEE conference on computer vision and pattern recognition}, pages 5285--5294, 2018.

\bibitem{multitaskoverview}
Sebastian Ruder.
\newblock An overview of multi-task learning in deep neural networks.
\newblock {\em arXiv preprint arXiv:1706.05098}, 2017.

\bibitem{multi13}
Kim-Han Thung and Chong-Yaw Wee.
\newblock A brief review on multi-task learning.
\newblock {\em Multimedia Tools and Applications}, 77:29705--29725, 2018.

\bibitem{transformer}
Ashish Vaswani, Noam Shazeer, Niki Parmar, Jakob Uszkoreit, Llion Jones, Aidan~N Gomez, {\L}ukasz Kaiser, and Illia Polosukhin.
\newblock Attention is all you need.
\newblock {\em Advances in neural information processing systems}, 30, 2017.

\bibitem{multi4}
Lin Xiao, Xin Huang, Boli Chen, and Liping Jing.
\newblock Label-specific document representation for multi-label text classification.
\newblock In {\em Proceedings of the 2019 conference on empirical methods in natural language processing and the 9th international joint conference on natural language processing (EMNLP-IJCNLP)}, pages 466--475, 2019.

\bibitem{multi12}
Tianhe Yu, Saurabh Kumar, Abhishek Gupta, Sergey Levine, Karol Hausman, and Chelsea Finn.
\newblock Gradient surgery for multi-task learning.
\newblock {\em Advances in Neural Information Processing Systems}, 33:5824--5836, 2020.

\bibitem{bigbird}
Manzil Zaheer, Guru Guruganesh, Kumar~Avinava Dubey, Joshua Ainslie, Chris Alberti, Santiago Ontanon, Philip Pham, Anirudh Ravula, Qifan Wang, Li~Yang, et~al.
\newblock Big bird: Transformers for longer sequences.
\newblock {\em Advances in neural information processing systems}, 33:17283--17297, 2020.

\end{thebibliography}

\end{document}